\documentclass[a4paper, 10pt, conference]{ieeeconf}      
\usepackage{FG2025}
\FGfinalcopy 

\IEEEoverridecommandlockouts                              
\usepackage{graphics} 
\usepackage{epsfig} 
\usepackage{mathptmx} 
\usepackage{times} 
\usepackage{amsmath} 
\usepackage{amssymb}  
\usepackage{amsfonts}
\usepackage{booktabs}
\usepackage{xcolor}

\def\etal{\textit{et al.}}

\usepackage{hyperref}

\usepackage{mathrsfs}

\usepackage{textcomp} 
\DeclareSymbolFont{textsymbols}{TS1}{\familydefault}{m}{n}
\SetSymbolFont{textsymbols}{bold}{TS1}{\familydefault}{m}{n}

\DeclareMathSymbol{\ulq}{\mathopen}{textsymbols}{39}
\DeclareMathSymbol{\urq}{\mathclose}{textsymbols}{39}
\DeclareMathAlphabet{\mathcal}{OMS}{cmsy}{m}{n}

\def\FGPaperID{0095} 

\title{\Large \bf
 Contrastive Language-Image Learning with Augmented Textual Prompts \\for 3D/4D FER Using Vision-Language Model
}

\author{\parbox{16cm}{\centering
    {\large Muzammil Behzad$^{1, *}\thanks{$^*$indicates corresponding author.}$, Guoying Zhao$^2$}\\
    {\normalsize
    $^1$ Information \& Computer Science Department, King Fahd University of Petroleum \& Minerals, Saudi Arabia \\
    $^2$ Center for Machine Vision and Signal Analysis, University of Oulu, Finland\\
    Email: \url{muzammil.behzad@kfupm.edu.sa}, \url{guoying.zhao@oulu.fi}\\
    }
    }
}

\begin{document}

\urlstyle{tt}

\ifFGfinal
\thispagestyle{empty}
\pagestyle{empty}
\else
\author{Anonymous FG2025 submission\\ Paper ID \FGPaperID \\}
\pagestyle{plain}
\fi
\maketitle

\begin{abstract}
In this paper, we introduce AffectVLM, a vision-language model designed to integrate multiviews for a semantically rich and visually comprehensive understanding of facial emotions from 3D/4D data. To effectively capture visual features, we propose a joint representation learning framework paired with a novel gradient-friendly loss function that accelerates model convergence towards optimal feature representation. Additionally, we introduce augmented textual prompts to enhance the model's linguistic capabilities and employ mixed view augmentation to expand the visual dataset. We also develop a Streamlit app for a real-time interactive inference and enable the model for distributed learning. Extensive experiments validate the superior performance of AffectVLM across multiple benchmarks.
\end{abstract}

\section{Introduction}
Facial Expression Recognition (FER) is a key research area within affective computing, focusing on analyzing and interpreting human emotions through facial analysis, with several applications in human-computer interaction \cite{chowdary2023deep}, mental health \cite{Foteinopoulou_2022}, education \cite{YADEGARIDEHKORDI2019103649}, and more \cite{7374704}. Given the complexity of the human emotions and the pioneering emotion theory proposed by Ekman \& Friesen \cite{ekman1971constants}, researchers have developed various models to identify and resolve potential research gaps \cite{LIU2023423}.


\begin{figure*}[t!]
    \centering
	\includegraphics[width=\linewidth]{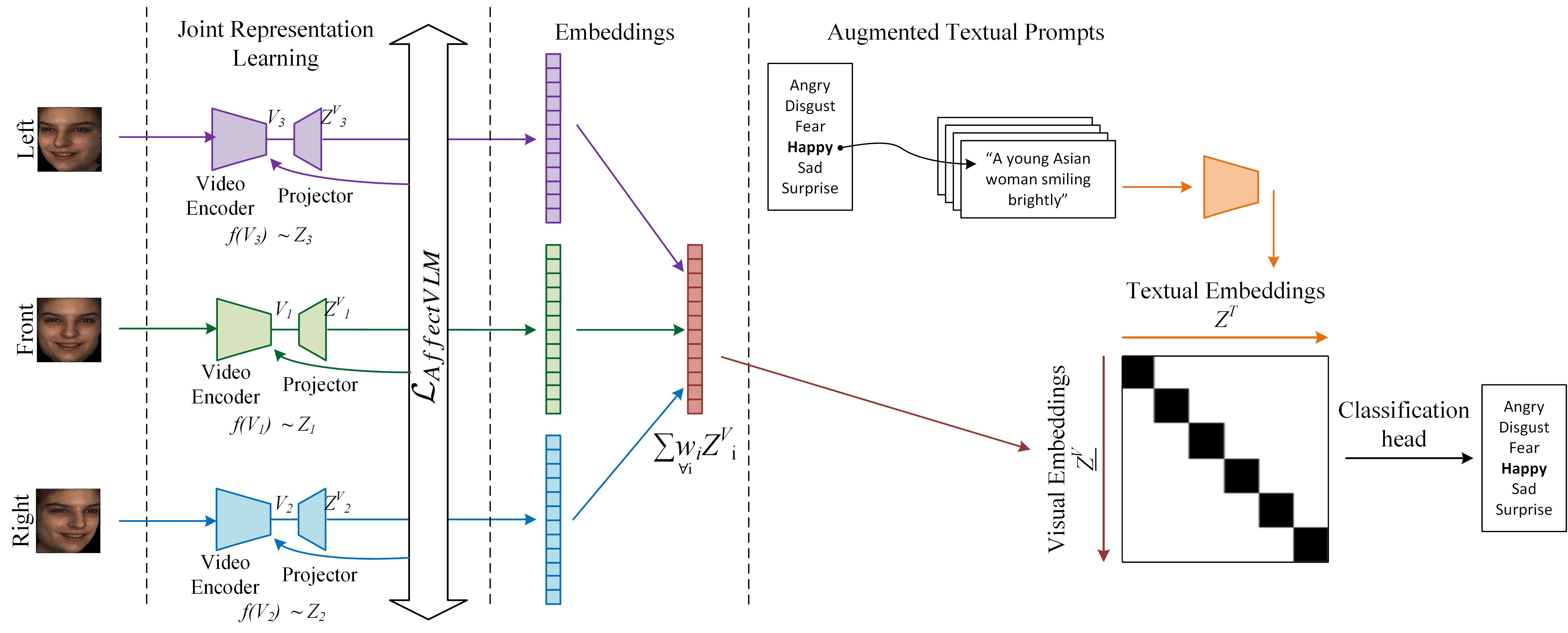}
	\caption{Overview of our proposed AffectVLM: Affective Vision-Language Model.}
	\label{MVTmodel}
\end{figure*}

The literature presents various methods to learn from the underlying 3D facial geometry. In this regard, the most widely used approaches include local feature-based methods~\cite{6460694, 5206613, li2015efficient}, template-based methods~\cite{4539275, 5597896}, curve-based methods~\cite{samir2009intrinsic, maalej2011shape}, and 2D projection-based methods~\cite{7944639, 8265585}. In recent years, 3D/4D FER has gained a significant attention as it enables deep learning models to extract additional discriminative facial features given the facial depth axis. For example, Yin~\etal~\cite{4813324} and Sun~\etal~\cite{Sun:2010:TVF:1820799.1820803} employed Hidden Markov Models (HMM) to capture temporal facial features from 4D facial scans. Likewise, Ben Amor~\etal~\cite{amor20144} demonstrated the effectiveness of a deformation vector field based on Riemannian analysis using a random forest classifier. Similarly, Sandbach~\etal~\cite{sandbach2012recognition} utilized Hidden Markov Models (HMM) and GentleBoost to learn free-form representations of 3D frames. Additionally, the authors in~\cite{FANG2012738} represented geometric coordinates and their normals as feature vectors, while another study~\cite{6130440} utilised dynamic local binary patterns (LBP) to recognize facial expressions using a support vector machine (SVM). In a related approach, the authors in~\cite{6553746} extracted features from polar angles and curvatures, proposing a spatio-temporal LBP-based feature extractor for recognition. 

Conversely, Li~\etal~\cite{8373807} introduced an intriguing framework for automatic 4D Facial Emotion Recognition (FER) using a dynamic geometrical image network. They generated geometrical images by calculating differential quantities from the given 3D facial point clouds. The emotion prediction is achieved through score-level fusion of the probability scores derived from various geometrical images.


\subsection{Motivations}
While effective, these traditional methods often rely on manually extracted features and localized cues, limiting performance and adaptability in the real-world scenarios. Inspired by the scaling victory of large language models~(LLMs) \cite{minaee2024largelanguagemodelssurvey} and large vision language models~(LVLMs)~\cite{zhang2024visionlanguagemodelsvisiontasks}, we work towards developing a VLM with joint representation learning capabilities to benefit from the underlying facial patterns stored as muscle movements in 3D/4D faces \cite{behzad2022deep, behzad2021self, behzad2020sparsity, behzad2020landmarks, behzad2021disentangling, behzad2019automatic, behzad2021towards, li20224dme}, for a significantly better facial expression recognition (FER). Specifically, instead of using conventional 2D faces only (e.g., \cite{8245803, 8844064, 9226082, 9369001}), this emotion recognition involves classifying facial expressions from 3D/4D faces with added spatio-temporal features, and the substantial results~\cite{li20113d, zhen2016muscular, 7163090, li2015efficient} have validated its advantages. VLMs offer a promising solution by learning joint representations, improving generalization across diverse datasets without manual feature engineering. This highlights the need for advanced approaches that leverage VLMs, which can learn from both visual and textual data~\cite{zhai2024finetuninglargevisionlanguagemodels}. By aligning visual cues with semantic labels, VLMs enhance the robustness of FER systems in varying conditions. However, existing 3D/4D datasets have limited capacity, which hinders the performance of deep learning models. This underscores the importance of improved augmentation strategies, such as spatial transformations, temporal variations, and multiview combinations, to address data scarcity and overfitting.

\subsection{Contributions}
To the best of our knowledge, existing literature lacks research on 3D/4D FER models utilizing VLMs. This gap arises primarily because implementing such models is not straightforward, accompanied by the complexity and variability in the 3D data structure. In this context, we introduce AffectVLM, a VLM based architecture \cite{radford2021learningtransferablevisualmodels}. The key features of our model are as follows:
\begin{enumerate}
	\item We propose a joint embedding space for coherent feature representation to integrate multiview embeddings.
     \item We also propose augmented textual prompts to extend the model's understanding of emotional semantics by augmenting the training labels with extended textual propmts.
    \item We introduce mixed view augmentation to generate diverse training samples by combining augmentations from multiple views, thereby enriching the dataset.
    \item We employ a novel gradient-friendly loss function for model's smoother convergence during training.
    \item We design AffectVLM with the capability to leverage distributed training for improved scalability.
    \item We develop a Streamlit application for user inference, enabling interactive access to the model for real-time emotion classification.
\end{enumerate}
Furthermore, it must be noted that AffectVLM is equally applicable to other downstream tasks like, face recognition, anti-spoofing and identity recognition.

\section{AffectVLM: Affective Vision-Language Model}
Our AffectVLM model's ability to incorporate multiviews offers a significant advantage with its robust and scalable implementation, making it highly effective and ready for deployment with minimal adjustments.

\subsection{Learning Joint Embedding Space Representations}
As shown in Fig. \ref{MVTmodel}, we tailor our model to learn joint embedding space to facilitate coherent feature representation across both visual and textual modalities. We use contrastive language-image pre-training (CLIP) \cite{radford2021learningtransferablevisualmodels} as a backbone VLM model and extend it for multiviews in our problem statement. Specifically, we aim to achieve joint embedding of visual features from multiple facial views and corresponding textual prompts. The model processes multiview data, thereby capturing frontal, left, and right perspectives, allowing it to learn a rich, holistic representation of emotions from different facial angles by optimizing the loss function jointly with the features from all views in the embedding space. Simultaneously, we integrate augmented textual prompts into the learning process as discussed in the next sub-section. This joint learning framework ensures that different views of the same emotion are aligned in a shared embedding space with their textual counterparts, promoting a deeper understanding of the emotion. This approach not only improves the model’s ability to generalize across unseen data but also enhances its robustness in recognizing emotions from varying viewpoints and contextual diversity. AffectVLM supports multiple model engines such as ViT(16/32)~\cite{dosovitskiy2021an}, and ResNet variants~\cite{he2016deep} to learn the underlying embeddings.
\begin{table}[b!]
    \centering
    \caption{Different textual prompts for ``Happy" expression}
    \begin{tabular}{c l}
        \toprule
        \textbf{Prompt ID} & \textbf{Textual Prompt} \\ 
        \midrule
        1 & ``\textit{A young woman with a joyful expression}" \\
        2 & ``\textit{An older man looking very happy}" \\
        3 & ``\textit{A smiling male full of joy}" \\
        4 & ``\textit{A young adult showing happiness}" \\
        5 & ``\textit{A middle-aged black woman looking very happy}" \\
        6 & ``\textit{Face of an older Asian woman showing a smile}" \\
        7 & ``\textit{A young Asian woman smiling brightly}" \\
        8 & ``\textit{An older Black female smiling}" \\ 
        \bottomrule
    \end{tabular}
    \label{tab:happy_prompts}
\end{table}

\subsection{Augmented Textual Prompts}
To enrich the semantic representation in our model's training process, we propose Augmented Textual Prompts. Considering the variability in expressions and subject-specific attributes like gender, age, ethnicity, and other contextual details, a single label such as ``Happy" may not capture the full spectrum of emotional states. By systematically generating multiple textual variations using predefined templates and adaptive modifications based on emotion and metadata, such as ``a joyful expression of a young female" or ``an older male smiling", we provide our VLM model with a broader linguistic context. This approach enhances the dataset with richer semantics and improves the model's ability to generalize across different data samples. The use of diverse prompts facilitates a deeper understanding of facial features. In Table \ref{tab:happy_prompts}, we present a list of textual prompts for the ``Happy" expression, generated using ChatGPT-3.5. Note that the range of prompts grows proportionally with more available annotations within the dataset. In this method, multiviews use different textual prompts during training to gain more semantic understanding.

\subsection{Mixed View Augmentation}
To address the challenge of limited data in 3D/4D FER~\cite{BEHZAD2021297}, we propose Mixed View Augmentation, an innovative method designed to enhance training data diversity by combining multiviews with existing common augmentation techniques such as flipping, rotation, cropping, and scaling. This approach allows us to apply different transformations across various views of the same emotion, creating a more extensive dataset. For instance, we can flip the frontal view of a face while cropping the left view, rather than augmenting solely within a single view. This cross-placing of augmentations significantly increases variability in the training set, facilitating more comprehensive learning of facial features. By employing this strategy, our model is better positioned to generalize across varying viewing conditions and emotional expressions, which ultimately enhances performance. This straightforward yet effective strategy not only reduces the risk of overfitting but also fosters robust learning, enabling the model to adapt to a wide array of visual contexts.

\subsection{Gradient-friendly Loss Function with Learnable Margin}
Using a loss function with different views can enhance the training process by enabling the model to learn more robust features across multiple perspectives. In this context, our model stands out due to its unique loss function that helps the network learn and converge more quickly. Unlike typical approaches, we work to optimize the loss together across all views. We define a shared embedding space $Z$ where features from different views are projected. Given $V_1, V_2, V_3$ as features from three different views, the shared embedding can be represented as:
\begin{equation}
    Z_i = f_{AffectVLM}(V_i),  i \in \{1,2,3\},
\end{equation}
where $f_{AffectVLM}$ is our model that projects features from each view into the shared space $Z$. We define our loss function as follows:
\begin{equation}
\label{lossfunction}
	\begin{aligned}
        \mathcal{L}_{AffectVLM} \triangleq \overbrace{ \underbrace{\sum_{i,j} 1-\Theta(Z_i,Z_j)}_{\text{for same emotion}} + \underbrace{\sum_{k,l} max(0,\Theta(Z_k,Z_l) - \alpha)}_{\text{for different emotion}} }^{\mathcal{L}_{mc}:\text{ multiview contrastive loss}} + 
		\\ 
		\underbrace{ max(0, || Z_{anchor}- Z_{positive}||_2^2 - || Z_{anchor}- Z_{negative}||_2^2 + \alpha)}_{{L}_{mt}:\text{ multiview triplet loss}},
	\end{aligned}
\end{equation}
where $\Theta(.)$ represents a similarity function such as Cosine similarity or Euclidean distance. Here, the multiview contrastive loss function $\mathcal{L}_{mc}$ is used to enforce similarity for the same emotion across views and dissimilarity for different emotions. Similarly, the multiview triplet loss $\mathcal{L}_{mt}$ is used to further enhance the learning process by explicitly constructing anchor, positive, and negative samples, where $Z_{anchor}$ is the shared representation from one view, $Z_{positive}$ is the representation of the same emotion from another view, and $Z_{negative}$
is from a different emotion. The margin $\alpha$ is a hyperparameter that helps ensure a gap between positive and negative pairs. In our setup, we initialize $\alpha$ as a learnable parameter, which allows the model to adapt the margin dynamically during training, ensuring model generalization while catering to the needs of various diverse datasets. 

\subsection{Performance Scalability with Distributed Leaning}
Our model's ability to train in a distributed environment is a significant feature that enhances its performance. We utilize PyTorch's distributed communication package (\url{torch.distributed}) alongside NVIDIA’s Collective Communications Library (NCCL) to efficiently distribute workloads across available resources. Our implementation dynamically determines the optimal training setup such as, multi-GPU, single-GPU, or CPU, based on resource availability, employing the \url{ddp} strategy in \url{pytorch-lightning} when multiple GPUs are present. To mitigate synchronization issues across GPUs or nodes, we dynamically allocate network \url{<port>} and \url{<IP>} addresses to establish flexible TCP communication sockets. This approach facilitates efficient multi-process parallelism, ensuring resource-efficient training of our AffectVLM model.

\subsection{Interactive Inference with Streamlit}
To make our model more user friendly, we have developed a \url{streamlit} application that serves as a service interface for real-time inference using our AffectVLM model. Users can upload multiple images for immediate processing, specifically three multiview images, enabling on-the-fly emotion classification based on the learned features from the joint embedding space. This architecture not only enhances user experience but also demonstrates the model’s scalability and adaptability to diverse input modalities in 3D/4D expression recognition tasks. Importantly, this user application opens directions for accelerated research development especially within real-time affective computing. We aim to publish this interactive application soon.


\section{Results and Discussions}
We validate our model using the Bosphorus \cite{savran2008bosphorus}, BU-3DFE \cite{yin20063d}, BU-4DFE \cite{4813324} and BP4D-Spontaneous \cite{ZHANG2014692} datasets. Consistent with prior works \cite{8373807, 8023848, 9320291, behzad2021Sparse3D}, we generate multiview 2D images from 3D/4D point clouds and apply rank pooling~\cite{bilen2018action} for video data to create dynamic images. A 10-fold subject-independent cross-validation is employed for all experiments.

\subsection{Performance on 3D FER}
Following existing protocols \cite{7944639, 8265585}, the BU-3DFE dataset with 101 subjects is divided into Subset I, which includes two higher intensity levels, and Subset II, which contains all four intensity levels. For the Bosphorus dataset, 65 subjects performed six expressions. Table~\ref{table:3DFERresults} shows our model's performance, achieving 92.63\% on Subset I, surpassing state-of-the-art results \cite{8265585}, and 89.74\% on the Bosphorus dataset. Notably, for Subset II, we outperform the current best method with an accuracy of 86.56\%, demonstrating our model's ability to effectively learn expressions.
\begin{table}[t!]
	\caption{Accuracy ($\%$) comparisons with state-of-the-art methods on the BU-3DFE Subset I and Subset II, and Bosphorus datasets.}
	\label{table:3DFERresults}
	\resizebox{\linewidth}{!}{%
		\begin{tabular}{l c}
			\hline
			Method & Subset I  (\color{blue}$\uparrow$\color{red}$\downarrow$\color{black})\\
			\hline			
			{\color{teal}Zhen \etal \cite{zhen2016muscular}} & 84.50 (\color{blue}8.13$\uparrow$\color{black}) \\ 
			{\color{teal}Yang \etal \cite{7163090}} & 84.80 (\color{blue}7.83$\uparrow$\color{black}) \\
			{\color{teal}Li \etal \cite{li2015efficient}} & 86.32 (\color{blue}6.31$\uparrow$\color{black}) \\
			{\color{teal}Li \etal \cite{7944639}} & 86.86 (\color{blue}5.77$\uparrow$\color{black}) \\ 
			{\color{teal}Oyedotun \etal \cite{8265585}} & 89.31  (\color{blue}3.32$\uparrow$\color{black}) \\ 
			\hline
			\textbf{{\color{magenta}AffectVLM (Ours)}} & \textbf{92.63}\\ 
			\hline
		\end{tabular}
		\begin{tabular}{l c c}
			\hline
			Method & Subset II (\color{blue}$\uparrow$\color{red}$\downarrow$\color{black}) & Bosphorus  (\color{blue}$\uparrow$\color{red}$\downarrow$\color{black})\\
			\hline
			{\color{teal}Li \etal \cite{li2015efficient}} & 80.42 (\color{blue}6.14$\uparrow$) & 79.72 (\color{blue}10.02$\uparrow$\color{black})\\ 
			{\color{teal}Yang \etal \cite{7163090}} & 80.46 (\color{blue}6.10$\uparrow$) & 77.50 (\color{blue}12.24$\uparrow$\color{black})\\ 
			{\color{teal}Li \etal \cite{7944639}} & 81.33 (\color{blue}5.23$\uparrow$) & 80.00 (\color{blue}9.74$\uparrow$\color{black})\\ 
			{\color{teal}Sui \etal \cite{sui2023afnet}} & - & 82.06 (\color{blue}7.68$\uparrow$\color{black})\\ 
			{\color{teal}Li \etal \cite{li2024drfer}} & - & 86.77 (\color{blue}2.97$\uparrow$\color{black})\\ 
			\hline
			\textbf{{\color{magenta}AffectVLM (Ours)}} & \textbf{86.56} & \textbf{89.74} \\ 
			\hline
		\end{tabular}
	}
\end{table}
\begin{table}[b!]
	\caption{Performance ($\%$) comparison of 4D FER with the state-of-the-art methods on the BU-4DFE dataset.}
	\label{table:4DFERresults}
	\begin{center}
		\begin{tabular}{l c c}
			\hline
			Method & Experimental Settings & Accuracy (\color{blue}$\uparrow$\color{red}$\downarrow$\color{black}) \\
			\hline
			{\color{teal}Sandbach \etal \cite{sandbach2012recognition}} & 6-CV, Sliding window & 64.60 ({\color{blue}34.76$\uparrow$})\\ 
			{\color{teal}Fang \etal \cite{6130440}} & 10-CV, Full sequence & 75.82 ({\color{blue}23.54$\uparrow$})\\ 
			{\color{teal}Xue \etal \cite{7045888}} & 10-CV, Full sequence & 78.80 ({\color{blue}20.56$\uparrow$})\\ 
			{\color{teal}Sun \etal \cite{Sun:2010:TVF:1820799.1820803}} & 10-CV, - & 83.70 ({\color{blue}15.66$\uparrow$})\\
			{\color{teal}Zhen \etal \cite{7457243}} & 10-CV, Full sequence & 87.06 ({\color{blue}12.30$\uparrow$})\\ 
			{\color{teal}Yao \etal \cite{10.1145/3131345}} & 10-CV, Key-frame & 87.61 ({\color{blue}11.75$\uparrow$})\\ 
			{\color{teal}Fang \etal \cite{FANG2012738}} & 10-CV, - & 91.00 ({\color{blue}8.36$\uparrow$})\\ 
			{\color{teal}Li \etal \cite{8373807}} & 10-CV, Full sequence & 92.22 ({\color{blue}7.14$\uparrow$})\\ 
			{\color{teal}Ben Amor \etal \cite{amor20144}} & 10-CV, Full sequence & 93.21 ({\color{blue}6.15$\uparrow$})\\ 
			{\color{teal}Zhen \etal \cite{8023848}} & 10-CV, Full sequence & 94.18 ({\color{blue}5.17$\uparrow$})\\ 
			{\color{teal}Bejaoui \etal \cite{Bejaoui2019}} & 10-CV, Full sequence & 94.20 ({\color{blue}5.15$\uparrow$})\\ 
			{\color{teal}Zhen \etal \cite{8023848}} & 10-CV, Key-frame & 95.13 ({\color{blue}4.23$\uparrow$})\\ 
			{\color{teal}Behzad \etal \cite{behzad2019automatic}}  & 10-CV, Full sequence & 96.50 ({\color{blue}2.86$\uparrow$})\\ 
			\hline
			\textbf{{\color{magenta}AffectVLM (Ours)}} & 10-CV, Full sequence & \textbf{99.36}\\ 
			\hline
		\end{tabular}
	\end{center}
\end{table}

\begin{table}[t!]
	\caption{Accuracy ($\%$) comparison on the BP4D-Spontaneous dataset.\vspace{0.1cm} \hspace{\textwidth} \hspace*{1.2cm}(a) Recognition \hspace{1cm} (b) Cross-Dataset Evaluation\vspace{-0.15cm}}
	\label{table:4DFERresults_part2}
	\resizebox{\linewidth}{!}{%
		\begin{tabular}{l c c}
			\hline
			Method & Accuracy (\color{blue}$\uparrow$\color{red}$\downarrow$\color{black})\\
			\hline
			{\color{teal}Yao \etal \cite{10.1145/3131345}} & 86.59 (\color{blue}5.99$\uparrow$\color{black})\\ 
			{\color{teal}Danelakis \etal \cite{danelakis2016effective}} & 88.56 (\color{blue}4.02$\uparrow$\color{black})\\ 
			\textbf{{\color{magenta}AffectVLM (Ours)}} & \textbf{92.58}\\ 
			\hline
		\end{tabular}
		\begin{tabular}{l c c}
			\hline
			Method & Accuracy (\color{blue}$\uparrow$\color{red}$\downarrow$\color{black})\\
			\hline
			{\color{teal}Zhang \etal \cite{ZHANG2014692}} & 71.00 (\color{blue}15.07$\uparrow$\color{black})\\ 
			{\color{teal}Zhen \etal \cite{zhen2017magnifying}} & 81.70 (\color{blue}4.37$\uparrow$\color{black})\\ 
			\textbf{{\color{magenta}AffectVLM (Ours)}} & \textbf{86.07}\\ 
			\hline
		\end{tabular}
	}
\end{table}

\subsection{Performance on 4D FER}
We conducted extensive experiments on the BU-4DFE dataset, which consists of posed video clips from 101 subjects with six facial expressions. As shown in Table~\ref{table:4DFERresults}, our model achieves a high accuracy of 99.36\%, outperforming all competing  methods by a significant margin. Compared to the top-performing state-of-the-art method~\cite{behzad2019automatic}, which achieved 96.50\%, our model outperforms it by \color{blue}2.86\%\color{black}, highlighting the effectiveness of our multiview architecture with its innovative loss function and augmentation strategies.

\subsection{Towards Spontaneous 4D FER}
We evaluated our model on the BP4D-Spontaneous dataset, which includes 41 subjects showing spontaneous expressions, including nervousness and pain. As shown in Table~\ref{table:4DFERresults_part2}, our method outperforms \cite{10.1145/3131345} by \color{blue}5.99\% \color{black} and \cite{danelakis2016effective} by \color{blue}4.02\% \color{black} in recognition accuracy. Additionally, we conducted cross-dataset evaluations~\cite{ZHANG2014692, zhen2017magnifying} by using BU-4DFE for training and BP4D-Spontaneous (Task 1 and Task 8) for validation, achieving an accuracy of 86.07\%. Our model surpasses \cite{ZHANG2014692} by \color{blue}15.07\% \color{black} and \cite{zhen2017magnifying} by \color{blue}4.37\% \color{black}, demonstrating the model's robustness and strong generalization to spontaneous expressions, making it well-suited for real-world applications.

\subsection{Performance Upgrade with Distributed Learning}
We use NVIDIA GeForce RTX 3090 Ti to demonstrate our model's efficiency on lower-end multi-GPU setups. As shown in Fig.~\ref{speedbar}, distributed training with 3 GPUs achieves average per-epoch times of 84s, 73s, and 50s for respective batches, reducing total training time to 11.67h, 10.13h, and 6.94h over 500 epochs, validating our model's scalability.

\subsection{Ablation Study}
\begin{figure}[b!]
    \centering
    \begin{minipage}{0.49\linewidth}
        \centering
        \includegraphics[width=\linewidth]{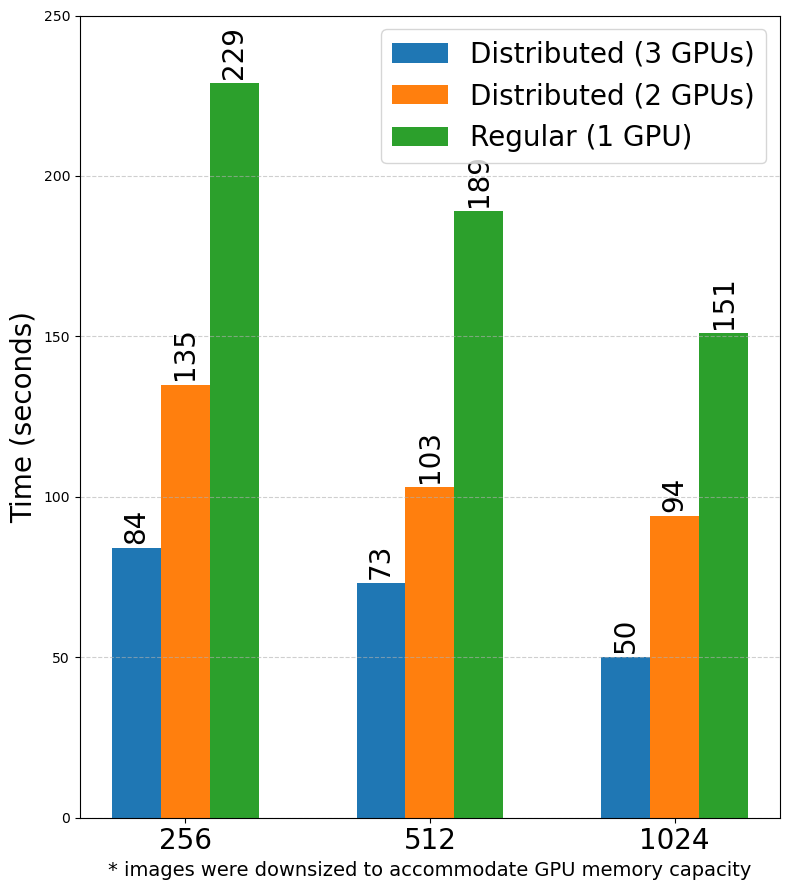}
        \caption{Performance comparisons of distributed learning.}
        \label{speedbar}
    \end{minipage}
    \hfill
    \begin{minipage}{0.49\linewidth}
        \centering
        \includegraphics[width=\linewidth]{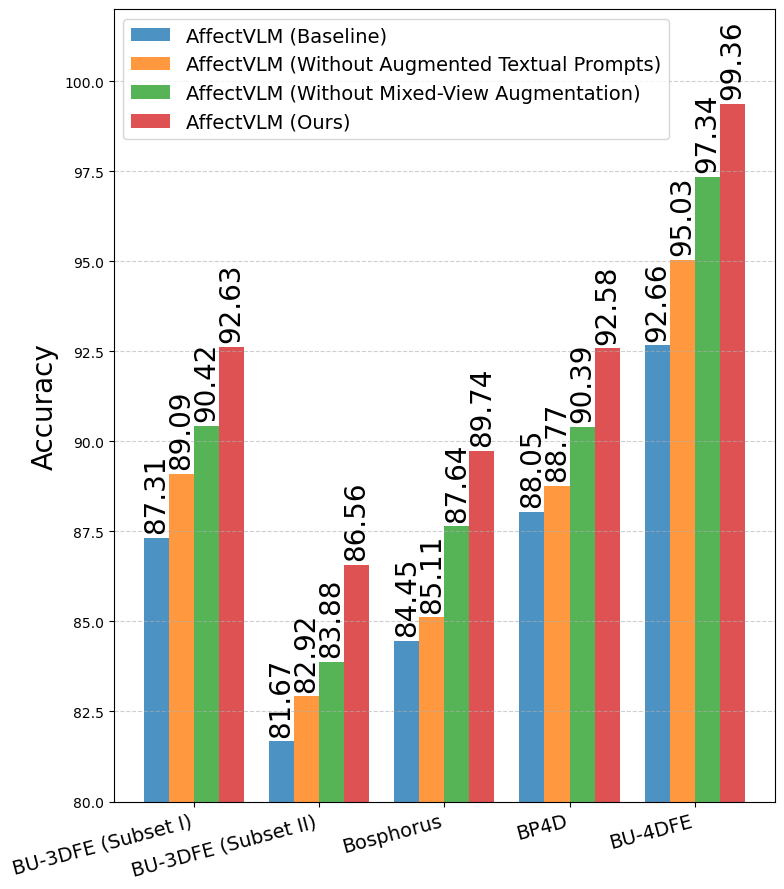}
        \caption{Ablation study showing the impact of individual components.}
        \label{ablation}
    \end{minipage}
\end{figure}

As shown in Fig.~\ref{ablation}, our ablation study demonstrates that augmented textual prompts and mixed-view augmentation significantly upgrades the performance, confirming their role in enhancing semantic understanding and generalization. Our AffectVLM model consistently achieves the highest accuracy, validating the effectiveness of our joint representation learning framework.

\section{Conclusion}
We presented AffectVLM, a vision-language model designed to integrate multiviews for a more comprehensive understanding of facial emotions from 3D/4D data. AffectVLM employed joint learning of feature representations from multiple views in a shared embedding space, optimizing them collectively. Additionally, we introduced augmented textual prompts to enhance the model's linguistic capabilities and a novel gradient-friendly loss function to improve convergence during training while using distributed learning. Extensive experiments validated the model's superior performance when compared to other state-of-the-art models.

\bibliographystyle{ieeetr}
\bibliography{main}

\end{document}